\documentclass[10pt,twocolumn,letterpaper]{article}

\usepackage[pagenumbers]{cvpr} %

\definecolor{cvprblue}{rgb}{0.21,0.49,0.74}
\usepackage[pagebackref,breaklinks,colorlinks,allcolors=cvprblue]{hyperref}

\def\paperID{*****} %
\def\confName{CVPR}
\def\confYear{2025}

\title{Owl-1: Omni World Model for Consistent Long Video Generation}

\author{Yuanhui Huang$^1$ \quad Wenzhao Zheng$^{1,}$\footnotemark[1] \quad Yuan Gao$^2$ \quad Xin Tao$^{2}$ \\ Pengfei Wan$^{2}$ \quad Di Zhang$^{2}$ \quad Jie Zhou$^1$ \quad Jiwen Lu$^1$ \\
$^1$Tsinghua University \quad $^2$Kuaishou Technology \\
\texttt{huangyh22@mails.tsinghua.edu.cn; wenzhao.zheng@outlook.com} 
}

\begin{document}

\twocolumn[{%
\renewcommand\twocolumn[1][]{#1}%
\vspace{-20mm}
\maketitle
\vspace{-12mm}
\begin{center}
    \centering
    \includegraphics[width=\linewidth]{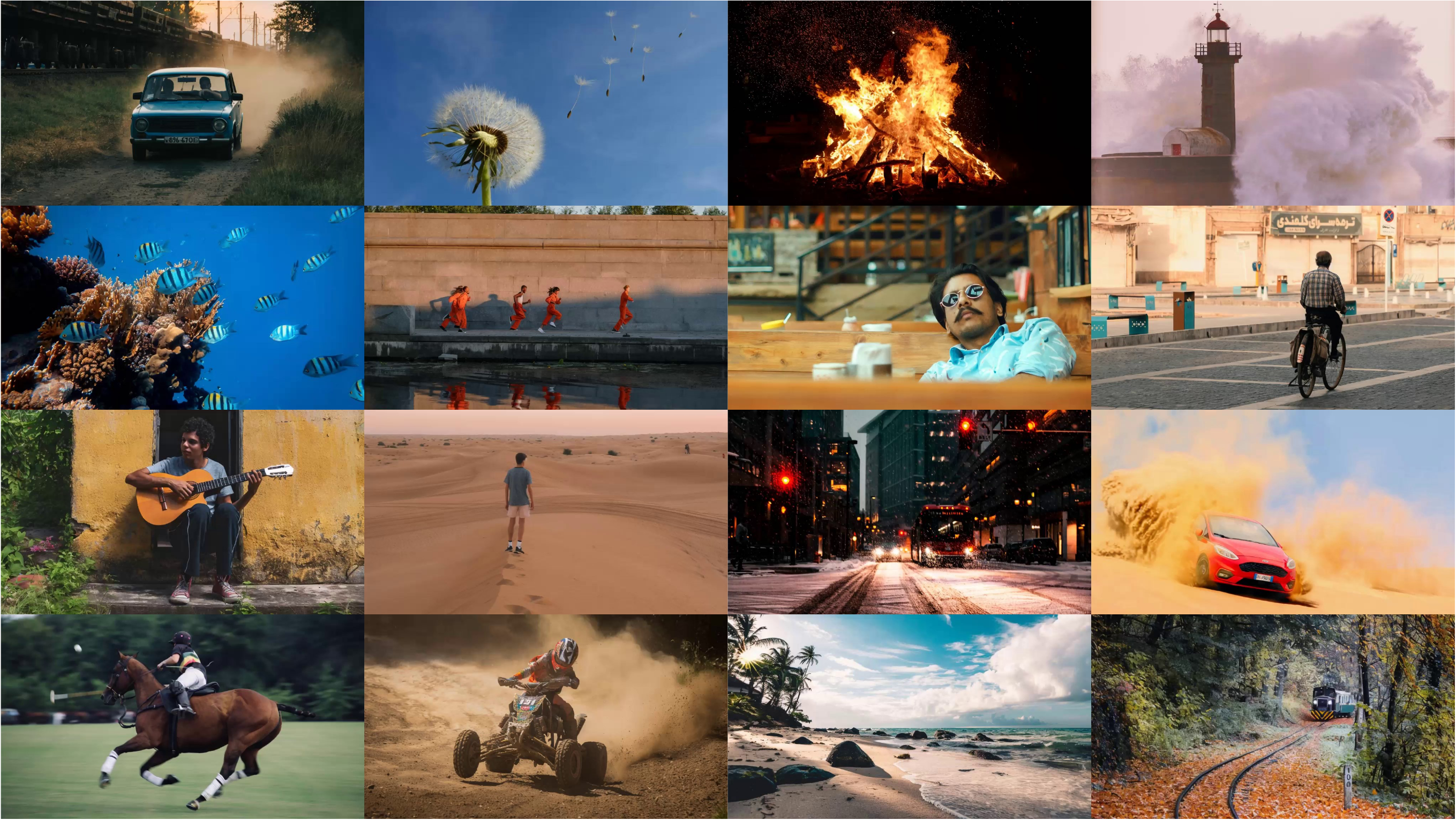}
    \vspace{-7mm}
    \captionof{figure}{
    Owl-1 approaches consistent long video generation with an omni world model, which models the evolution of the underlying world with latent state, explicit observation and world dynamics variables.
    }
    \vspace{-1mm}
\label{teaser}
\end{center}%
}]

\renewcommand{\thefootnote}{\fnsymbol{footnote}}
\footnotetext[1]{Project leader.}
\renewcommand{\thefootnote}{\arabic{footnote}}

\begin{abstract}
Video generation models (VGMs) have received extensive attention recently and serve as promising candidates for general-purpose large vision models.
While they can only generate short videos each time, existing methods achieve long video generation by iteratively calling the VGMs, using the last-frame output as the condition for the next-round generation.
However, the last frame only contains short-term fine-grained information about the scene, resulting in inconsistency in the long horizon. 
To address this, we propose an Omni World modeL (Owl-1) to produce long-term coherent and comprehensive conditions for consistent long video generation. 
As videos are observations of the underlying evolving world, we propose to model the long-term developments in a latent space and use VGMs to film them into videos.
Specifically, we represent the world with a latent state variable which can be decoded into explicit video observations.
These observations serve as a basis for anticipating temporal dynamics which in turn update the state variable.
The interaction between evolving dynamics and persistent state enhances the diversity and consistency of the long videos.
Extensive experiments show that Owl-1 achieves comparable performance with SOTA methods on VBench-I2V and VBench-Long, validating its ability to generate high-quality video observations.
Code: \url{https://github.com/huang-yh/Owl}.

\end{abstract}
 
\vspace{-6mm}
\section{Introduction}
\label{sec:intro}

With the success of image generative models~\cite{rombach2022high,schuhmann2022laion,zheng2024cogview3,podell2023sdxl,guo2023animatediff,betker2023improving}, video generation~\cite{ho2022imagen,hong2022cogvideo,huang2023vbench,singer2022make,villegas2022phenaki} have also garnered increasing attention.
While existing video generation models (VGMs)~\cite{chen2024videocrafter2,ren2024consisti2v,blattmann2023stable,xing2025dynamicrafter} have achieved commercial-grade performance, the durations of videos are still short.
The long video generation methods~\cite{ju2024miradata,opensora,opensoraplan,yang2024cogvideox,wang2023lavie} remedies this issue by focusing on improving the length and consistency of generated videos, facilitating a variety of newly rising tasks such as video extension~\cite{xing2025dynamicrafter}, film generation~\cite{zhao2024moviedreamer} and world simulation~\cite{qin2024worldsimbench}.

Despite the promising applications, how to increase the video length while preserving consistency remains an open question.
Several work~\cite{bain2021frozen,opensora} investigates the 3D variational autoencoder (VAE) which compresses a video in both spatial and temporal dimensions in order to generate long videos in a single denoising process of a latent diffusion model.
Although the video consistency is inherently guaranteed in the diffusion process, the length of the generated videos is limited by the computational resources end further expanding the video length requires retraining the diffusion model.
Another line of work approaches long video generation through divide-and-conquer, which first generates the key frames of a long video and then interpolates between successive key frames~\cite{yin2023nuwa,ge2022long}.
However, these methods are dependent on the duration of the training video data, thus lacking scalability.
In addition, iteratively prompting a video diffusion model for short clip generation is also a promising paradigm to generate long videos~\cite{henschel2024streamingt2v,chen2023seine,villegas2022phenaki}.
To achieve consistency, these approaches design their prompts based on historical clips and texts in each iteration.
Nonetheless, current practices for prompt construction usually take the last frames of the direct adjacent clip, which only contain short-term information about the scene, resulting in inconsistency in the long horizon.

In this paper, we propose an Omni World modeL (Owl-1) to produce long-term coherent and comprehensive conditions for consistent long video generation.
Since videos are observations of the underlying evolving world, which establishes the temporal consistency of videos, we propose to model the long-term developments in a latent space and use VGMs to film them into videos. 
To elaborate, we represent the world with a latent state variable which encodes both the current and historical information about the underlying world.
Similar to the filming process, the state variable decodes into video clips with VGMs as observations of the world.
Based on these observations, we further anticipate future world dynamics which drive the evolution of the world and update the latent state variable.
Up to now, we have constructed an autoregressive state-observation-dynamics model to simulate the closed-loop evolution of the world, which improves the coherence of long videos with the consistent latent states, and enhances the content diversity with dynamics predictions.
To effectively model the relationship of these three components, we employ a pretrained large multimodal model (LMM) to take advantage of its general reasoning ability.
Additionally, we adopt a video diffusion model to decode latent states into short video clips.
Owl-1 achieves comparable performance with SOTA methods on VBench-I2V and VBench-Long, validating its ability to generate high-quality video observations. 

\begin{figure}[t]
\centering
\includegraphics[width=0.47\textwidth]{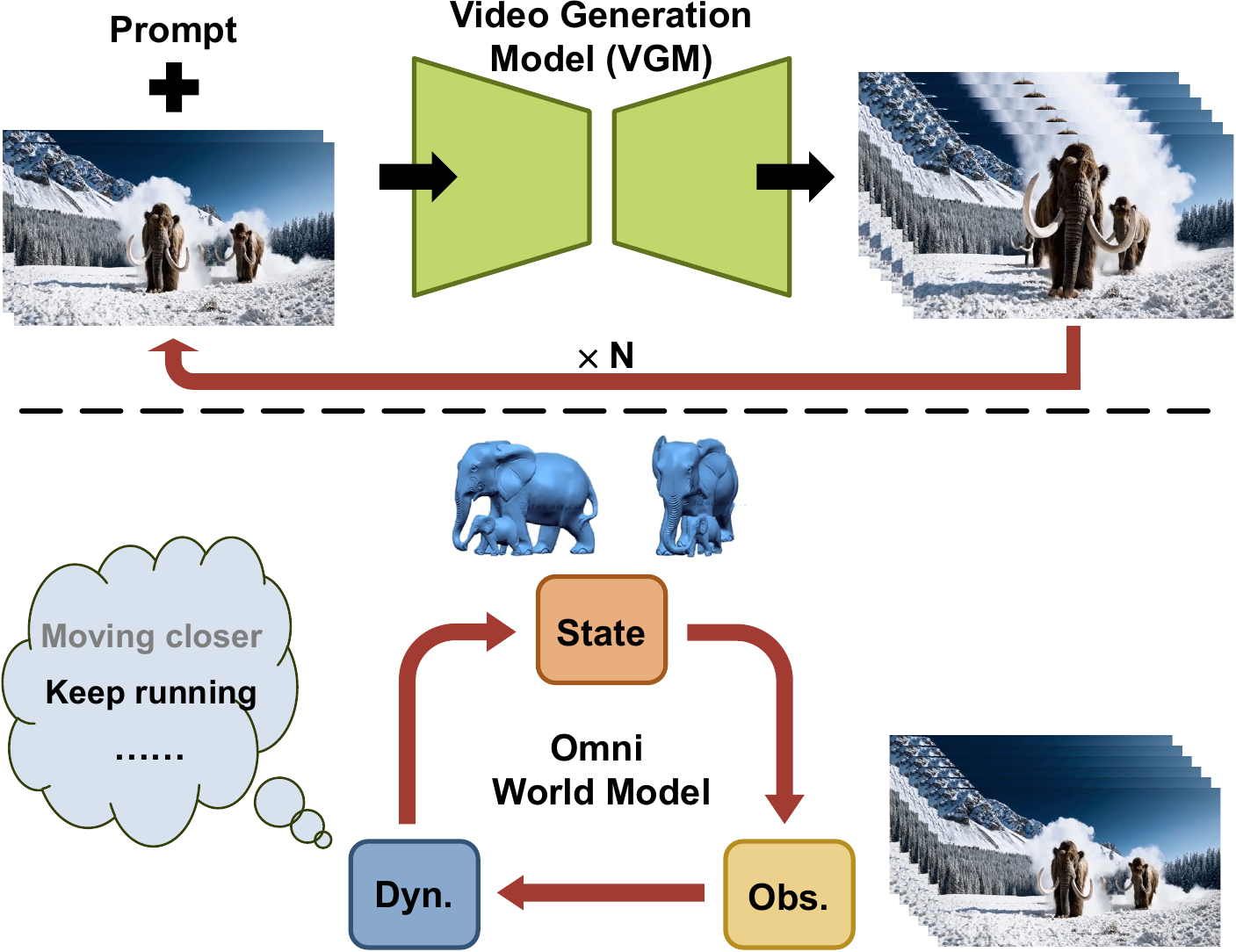}
\vspace{-1mm}
\caption{\textbf{Iterative long video generation.}
Conventional iterative long video generation methods use the last-frame output as the condition for the next-round generation, which lacks long-term consistency.
Our method constructs an omni world model for comprehensive conditioning.
}
\label{fig:motivation}
\vspace{-6mm}
\end{figure}

\section{Related Work}
\label{sec: related work}

\begin{figure*}[t]
\centering
\includegraphics[width=\textwidth]{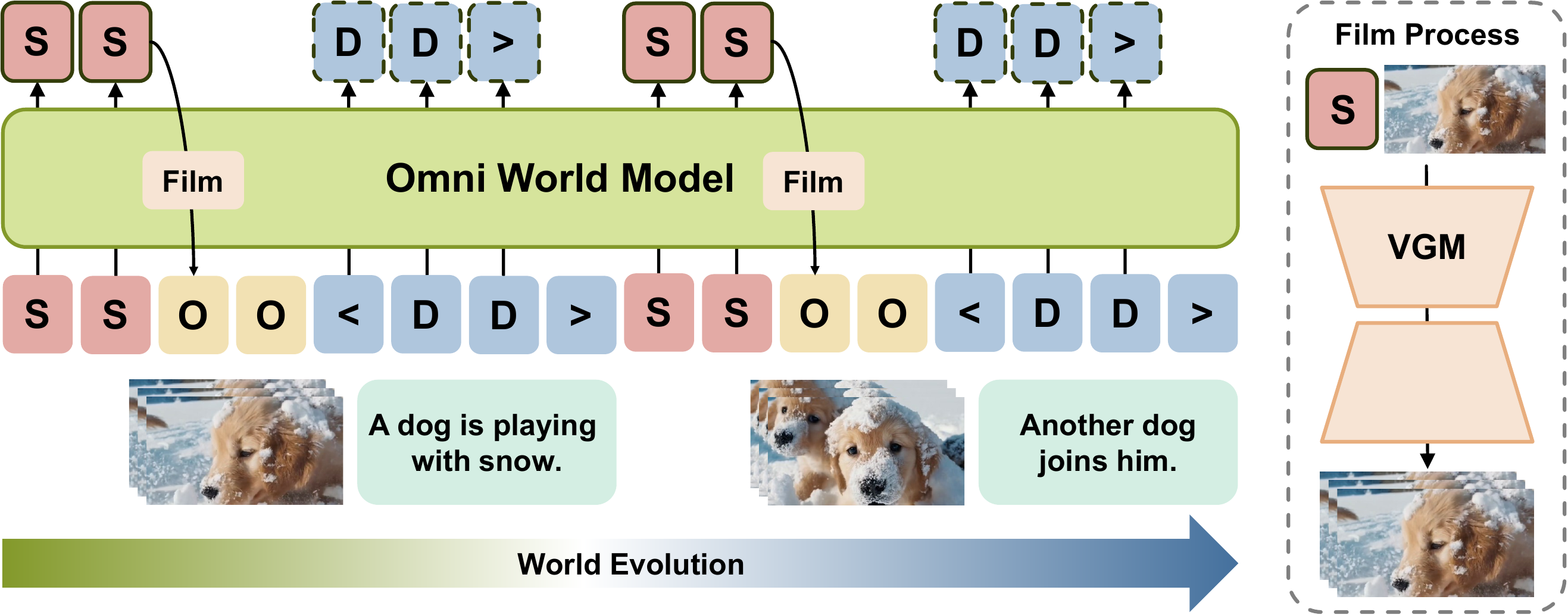}
\vspace{-4mm}
\caption{\textbf{Overall framework.}
Our Owl-1 models the evolution of the world with the latent state variables $\mathbf{s}$, and film them into video observations $\mathbf{o}$ along the generation process.
We also incorporate anticipation of the world dynamics $\mathbf{d}$ to explicitly drive the evolution.
}
\label{fig:pipeline}
\vspace{-5mm}
\end{figure*}

\textbf{Short video generation.}
In the realm of computer vision, video generation has emerged as a pivotal area of research, garnering significant attention due to its broad applications.
Short video generation investigates how to generate videos based on text (and/or image) conditions, where the alignment between the generated video and the given conditions is one of the primary evaluation criteria.
For text conditions, most methods~\cite{blattmann2023stable,xing2025dynamicrafter,hong2022cogvideo} encode them with pretrained text encoders~\cite{ni2021sentence,radford2021learning}, and incorporate the textual features using cross attention.
In addition, image-to-video models requires the generated video to incorporate the specified image conditions.
In order to effectively fuse the fine-grained visual information, several approaches~\cite{opensora,henschel2024streamingt2v} directly replace or concatenate the diffusion features with the encoded features of the image condition.
Other methods~\cite{xing2025dynamicrafter} also transform the image condition into tokens similar to the textual features, and apply cross attention between the diffusion features and image tokens to preserve coarser level of details such as visual styles and background.
Our Owl-1 uses both the latent state and optional image conditions from the last clip for consistent and smooth generation of the next clip.

\textbf{Long video generation.} 
As an important extension of the application scope of video generation models, long video generation focuses on improving the length and consistency of generated videos.
To achieve this, several work attempts to enhance the video durations in a single generation process, by designing 3D VAEs that are able to compress longer videos~\cite{opensora,bain2021frozen} or investigating the temporal modules in VGMs for efficient generation~\cite{xing2025dynamicrafter}.
Although the end-to-end generation pipeline inherently guarantees the video consistency, the length of generated videos is constrained by limited computational resources.
To remedy this issue, the divide-and-conquer approach simplifies the task by first identifying key frames that outline the main narrative and then generating the intervening frames to create a cohesive long video.
However, these methods are dependent on training video data of long durations which are still insufficient, thus lacking scalability.

On the other hand, the temporal autoregressive paradigm adopts a sequential approach to generate short video segments based on prior conditions. 
Within this paradigm, various models have been employed, including diffusion models~\cite{henschel2024streamingt2v,villegas2022phenaki}, spatial autoregressive models~\cite{kondratyuk2023videopoet}, and GAN models~\cite{skorokhodov2022stylegan}.
The key challenge here is to ensure the consistency between temporally distant clips to achieve coherent long video generation. 
Most work directly uses the last frames of the previous generated clip as visual clues for the next-round generation, which only contain short-term information about the scene, resulting in a limited temporal receptive field and inconsistency in the long horizon.
In contrast, our Owl-1 employs the latent state variable which encodes both the current and historical information about the underlying world to achieve extensive temporal receptive field and video consistency.

\textbf{Video generation world models.}
Video generation models are promising candidates for world models~\cite{zhu2024sora} which aims to model the evolution of the environment.
For videos of short durations, the generated content may reflect certain physical laws~\cite{liu2024sora}, indicating that the video generation model has learned some general knowledge about the world.
For a longer horizon, the emphasis of video generation world models lies in capturing overall dynamics that drive environment evolution~\cite{zhu2024sora}.
Although such models have been proposed in autonomous driving~\cite{wang2024occsora,zheng2025occworld} and embodied intelligence~\cite{brohan2023rt}, they can only predict structured actions instead of general world dynamics in the form of natural language.
As for general video generation, most existing methods focus on improving the alignment of generated videos and given text conditions, lacking the ability to anticipate the world dynamics.
In addition to conditional video generation, our Owl-1 is capable of predicting future dynamics to generate long videos with diverse content.

\section{Proposed Approach}

\label{sec: approach}
In this section, we present our method of omni world model for consistent long video generation.
To formulate this task mathematically, we aim to generate a long video consisting of a sequence of video clips $\mathbf{v} = \{..., \mathbf{o}_{t-1}, \mathbf{o}_t, \mathbf{o}_{t+1}, ...\}$ given a starting image $\mathbf{I}$ and a text description $\mathbf{d}_0$ as input.

\subsection{Omni World Model}
\label{subsec: owl}
Videos are fundamentally recorded observations of the underlying evolving world, whose long-term consistency is inherently guaranteed in the coherence of the world itself.
Therefore, maintaining consistency in long videos from the perspective of the implicit world is a more reasonable and essential approach, compared with the explicit pixel-space methods.
However, the real world constitutes a complex high-dimensional system, and the cost of directly modeling such a system is unacceptable.
Inspired by the world models in the field of embodied intelligence~\cite{brohan2022rt1}, we represent the world using a set of latent state variables $\{..., \mathbf{s}_{t-1}, \mathbf{s}_t, ...\}$.
Each state $\mathbf{s}_t$ not only encodes information about the world at the current moment $t$, but also incorporates historical information about the evolution of the world, i.e. $\{..., \mathbf{s}_{t-2}, \mathbf{s}_{t-1}\}$.
Since state variables serve as an purely implicit representation of the world, we introduce a state decoder $\mathcal{D}$ to obtain explicit video observations $\{..., \mathbf{o}_{t-1}, \mathbf{o}_t, ...\}$ from the state variables:
\begin{equation}
    \mathbf{o}_t = \mathcal{D}(\mathbf{s}_t, \mathbf{o}_{t-1}),
    \label{eq: observation predict}
\end{equation}
where we incorporate the last observation $\mathbf{o}_{t-1}$ to ensure short-term fine-grained smoothness of successive observations, while the current state $\mathbf{s}_t$ is primarily responsible for the long-term consistency. 

Most work approaches long video generation in the same way as short video generation, which overlooks the variation of video content in a long horizon, resulting in repeated generation of homogeneous content.
In our omni world world, we explicitly take the world dynamics $\{..., \mathbf{d}_{t-1}, \mathbf{d}_{t}, ...\}$ into consideration which drives the evolution of the underlying world and takes the form of texts.
To elaborate, we predict the current world dynamics $\mathbf{d}_t$ from state variables and video observations:
\begin{equation}
    \mathbf{d}_t = f(\mathbf{s}_t, \mathbf{o}_t),
    \label{eq: dynamics predict}
\end{equation}
where $f(\cdot)$ denote the world dynamics prediction function.
Furthermore, current world dynamics, in turn, updates the state variable, advancing the evolution of the world:
\begin{equation}
    \mathbf{s}_{t+1} = g(\mathbf{s}_t, \mathbf{d}_t),
    \label{eq: state predict}
\end{equation}
where $g(\cdot)$ represents the world state prediction function.
With Eq.~(\ref{eq: observation predict})(\ref{eq: dynamics predict})(\ref{eq: state predict}), we have constructed a state-observation-dynamics triplet to simulate the evolution of the world and also obtain consistent video clips along the evolution, as in Figure~\ref{fig:pipeline}. 
This formulation improves the consistency of long videos by modeling the underlying world and generates diverse video clips through explicit dynamics prediction.

\subsection{Comprehensive Condition from Latent State}
\label{subsec: summary of history}
The key challenge in the temporal autoregressive paradigm for long video generation lies in the design of the condition used for generating the next clip.
Most existing methods directly take the last frames of the previous clip as condition, which only considers the short-term smoothness between consecutive clips, and overlooks consistency issues in the long-term such as style, character identity, background etc.
Our Owl-1 takes the latent state variable as a comprehensive condition for the long-term consistency, because the derivation of the current state $\mathbf{s}_t$ inherently includes the information of all previous observations: %
\begin{equation}
    \mathbf{s}_{t+1} = g(\mathbf{s}_t, f(\mathbf{s}_t, \mathbf{o}_t)) = h(\mathbf{s}_0, \mathbf{o}_0, ..., \mathbf{o}_{t-1}, \mathbf{o}_t),
\end{equation}
which is derived by plugging Eq.~(\ref{eq: dynamics predict}) into Eq.~(\ref{eq: state predict}) and iteratively replacing $\mathbf{s}_t$ with $\mathbf{s}_{t-1}$ and $\mathbf{o}_{t-1}$.

For implementation of our Owl-1, we take advantage of a large multimodal model (LMM) to instantiate the functions $f(\cdot)$ and $g(\cdot)$, in order to take advantage of its common knowledge from the large scale pretraining on textual and visual data, and its large receptive field as well.
And we instantiate the state decoder $\mathcal{D}$ with a pretrained video diffusion model for their capability to generate short videos of high quality.
To incorporate the state-observation-dynamics triplet into the framework of LMM, we design the format of the input and output sequences:
\begin{equation}
    {\rm Seq} = \big[..., \mathbf{s}_t, \mathbf{o}_t, \mathbf{d}_t, ...\big],
    \label{eq: state summarize}
\end{equation}
where we iteratively feed the basic triplet $(\mathbf{s}_t, \mathbf{o}_t, \mathbf{d}_t)$ into the LMM.
For latent state $\mathbf{s}_t$, we use a set of $Q$ learnable query embeddings as input tokens to LMM, since it has no ground truth and thus cannot be quantized into discrete tokens.
As for video observation $\mathbf{o}_t$, we uniformly sample a number of key frames from the current clip, and use the pretrained vector-quantized variational autoencoder (VQVAE) of the LMM to transform the key frames into visual tokens.
Moreover, we directly use the text tokenizer of the LMM to convert the textual world dynamics $\mathbf{d}_t$ into discrete input tokens.
In summary, we use the LMM to model the closed-loop state-observation-dynamics evolution, where the state variable aggregates the information of all previous video observations (Eq.~(\ref{eq: state summarize})) and serves as comprehensive condition for the next-round generation (Eq.~(\ref{eq: observation predict})).

\subsection{Anticipation of Future Dynamics}
\label{subsec: anticipate future}
In the context of long video generation, anticipating future dynamics is crucial for maintaining consistency and coherence across extended video sequences. 
Our Owl-1 predicts and integrates these future dynamics $\mathbf{d}_t$ into the evolution of the latent state $\mathbf{s}$, thereby enriching the content diversity and ensuring temporal consistency in the generated videos.
As indicated by Eq.~(\ref{eq: dynamics predict}), the prediction of world dynamics $\mathbf{d}_t$ relies on the current video observation $\mathbf{o}_t$ as short-term reference, and the current latent state $\mathbf{s}_t$ as source of long-term information.
Once the current dynamics $\mathbf{d}_t$ is predicted, we integrate it into the latent state variable $\mathbf{s}_t$ to update the world state for the next-round generation.
To train the dynamics anticipation ability of the LMM, we adopt the next-token prediction paradigm and use the textual ground truth dynamics for teacher-forcing supervision.

The anticipation of future dynamics is important for content diversity of generated long videos and world modeling.
By enabling the anticipation of subsequent events within a video sequence, our Owl-1 enhances the richness of the generated content, moving beyond repeated generation of homogeneous content to capturing the essence of dynamic scenarios. 
Furthermore, future dynamics prediction serves as a cornerstone for constructing plausible world models, which are instrumental in simulating and understanding complex environments. 
Our Owl-1 not only predicts real-world behaviors but also allow for the incorporation of control mechanisms by replacing the anticipated dynamics with user-input control signals, facilitating the generation of content that is not only predictable but also controllable. 

\begin{table*}[!ht]
\small
\centering
\caption{\textbf{Evaluation results on VBench-I2V.} Subj., Bkgd. and Consist. denote Subject, Background and Consistency, respectively. \textbf{Bold:} best results. \underline{Underline}: second best.
Our Owl-1 achieves comparable performance with state-of-the-art image-to-video models.
}
\vspace{-2mm}
\tabcolsep=0.15cm
\def\arraystretch{1.0}
\setlength\tabcolsep{2.1 pt}
\begin{tabular}{@{}lcccccccccc@{}}
\toprule
    Method  &  \begin{tabular}[c]{@{}c@{}}Video-Image\\ Subj. Consist.\end{tabular} %
    & \begin{tabular}[c]{@{}c@{}}Video-Image\\ Bkgd. Consist.\end{tabular} %
    & \begin{tabular}[c]{@{}c@{}}Subject\\ Consist.\end{tabular} %
    & \begin{tabular}[c]{@{}c@{}}Bkgd.\\ Consist.\end{tabular} %
    & \begin{tabular}[c]{@{}c@{}}Motion\\ Smoothness\end{tabular} %
    &  \begin{tabular}[c]{@{}c@{}}Dynamic\\ Degree\end{tabular} %
    &  \begin{tabular}[c]{@{}c@{}}Aesthetic\\ Quality\end{tabular} %
    &  \begin{tabular}[c]{@{}c@{}}Imaging\\ Quality\end{tabular} %
    &  \begin{tabular}[c]{@{}c@{}}Temporal \\ Flickering\end{tabular} %
    &  \begin{tabular}[c]{@{}c@{}}Total \\ Score\end{tabular} %
    \\ 
 \midrule
VideoCrafter-I2V~\cite{chen2024videocrafter2} & 91.17& 91.31 &\underline{97.86} &\textbf{98.79} &98.00 &22.60 &60.78 &\underline{71.68} &98.19 &85.14 \\
ConsistI2V~\cite{ren2024consisti2v} &95.82 &95.95 &95.27 &98.28 &97.38 &18.62 &59.00 &66.92 &97.56 &86.84 \\
SEINE-512x512~\cite{chen2023seine} &97.15 &96.94 &95.28 &97.12 &97.12 &27.07 &64.55 &71.39 &97.31 &88.42 \\
I2VGen-XL~\cite{zhang2023i2vgen} &96.48 &96.83 &94.18 &97.09 &98.34 &26.10 &64.82 &69.14 &98.58 &88.48 \\
Animate-Anything~\cite{dai2023animateanything} &\textbf{98.76} &\underline{98.58} & \textbf{98.90} &98.19 &\underline{98.61} &02.68 &\textbf{67.12} &\textbf{72.09} &98.14 &89.76 \\
SVD-XT-1.0~\cite{blattmann2023stable} &97.52 &97.63 &95.52 &96.61 &98.09 &\textbf{52.36} &60.15 &69.80 &\textbf{99.09} & \underline{89.87} \\
DynamiCrafter-1024~\cite{xing2025dynamicrafter} &\underline{98.17} & \textbf{98.60} &95.69 &97.38 &97.38 &\underline{47.40} &\underline{66.46} &69.34 &97.63 &\textbf{90.25} \\
\midrule
\textbf{Owl-1} &  97.40 & 97.29 & 97.28 & \underline{98.54} & \textbf{98.92} & 21.63  & 61.89  &  69.66 & \underline{98.69} & 89.15 \\ \bottomrule
\end{tabular}
\label{tab:i2vbench}
\vspace{-5mm}
\end{table*}

\subsection{Multi-Stage Training}
\label{subsec: multistage train}

Several challenges exist in the training process of our Owl-1:
1) Since the LMM and the video diffusion model are separately pretrained, it is nontrivial to align these two models.
2) Our Owl-1 is designed for long-term world modeling, which requires video data with long duration and dense captions. 
However, given the scarcity of such high-quality data, it would be infeasible to train these large models with billions of parameters directly for the purpose of world model.
Therefore, we carefully design a multi-stage training scheme for our Owl-1 which consists of alignment, generative pretraining and world model training.

\textbf{The alignment stage} primarily enforces the consistency between the state variables $\mathbf{s}_t$ from the LMM and the textual conditions of the video diffusion model, which serves as a good initialization for the subsequent generative pretraining stage.
Specifically, we freeze the video diffusion model to preserve its ability of generating short videos and only trains the LMM at this stage.
We use general datasets for video generation in this stage, which provide videos of varying lengths and one single description for each video.
For each sample $(\mathbf{v}, \mathbf{t})$, we first segment the video into short clips of fixed length $\mathbf{v} = \{..., \mathbf{o}_{t-1}, \mathbf{o}_t, \mathbf{o}_{t+1}, ...\}$, and construct the input sequence as:
\begin{equation}
    {\rm Seq}_{align} = \big[\mathbf{I}, \mathbf{t}, \mathbf{s}_0, \mathbf{o}_0, \mathbf{t}, ..., \mathbf{s}_t, \mathbf{o}_t, \mathbf{t}, ...\big],
    \label{eq: align seq}
\end{equation}
where $\mathbf{I}$ represents the first frame, and we use the same text dynamics $\mathbf{t}$ for every triplet since the general video generation datasets do not provide dense captions for every clip and the content of the video remains largely unchanged throughout its duration.
To train the LMM to align with the textual conditions of video diffusion model, we minimize the L2 distance between the latent state $\mathbf{s}_t$ and the text features from the text encoder of video diffusion model $\mathcal{T}$:
\begin{equation}
    \mathcal{L}_{align} = {\rm MSE}(\mathbf{s}_t, \mathcal{T}(\mathbf{t})).
    \label{eq: loss align}
\end{equation}
The alignment stage enforces the consistency between the state variable and the textual conditions of the video diffusion model, which is pivotal for the stability of subsequent training given the distinction between the LMM and the video diffusion model.

\textbf{The generative pretraining stage} finetunes the LMM and the video diffusion model in a joint manner, to train the ability of the video diffusion model as the state decoder (Eq.~\ref{eq: observation predict}), which translates the latent state $\mathbf{s}_t$ into explicit video observations $\mathbf{o}_t$.
We adopt the same general video generation datasets and thus the same input sequence in Eq.~(\ref{eq: align seq}) for this stage.
Since the purpose of the ${\rm MSE}$ loss in the alignment stage (Eq.~(\ref{eq: loss align})) is only to provide an initialization, we discard it in the generative pretraining stage and substitute the latent state $\mathbf{s}_t$ for the original text condition of the video diffusion model.
We supervise these two models with only the denoising target of diffusion models:
\begin{equation}
    \mathcal{L}_{pretrain} = ||\mathbf{\epsilon} - \hat{\epsilon}_\mathcal{D}(\mathbf{o}_{t,m}, m, \mathbf{s}_t, \mathbf{o}_{t-1}) ||_2^2,
    \label{eq: denoise target}
\end{equation}
where $m$, $\mathbf{o}_{t,m}$ represent the denoising timestamp and the noisy video observation, respectively.
By training the video diffusion model with the latent state $\mathbf{s}_t$ as conditional input, we turn the video diffusion model into a photographer who films the latent world into explicit videos.

\textbf{The world model training stage} mainly incorporates the prediction of world dynamics $\mathbf{d}_t$ into our Owl-1.
It is based on the large scale pretraining of the second stage, which unifies the LMM and the video diffusion model as a preliminary Owl-1 capable of generating latent states $\mathbf{s}_t$ as comprehensive conditions for video clip generation.
Now we further finetune the LMM and video diffusion model on a small amount of video data with longer duration and dense captions due to its scarcity.
To achieve this, we change the input sequence of the LMM as:
\begin{equation}
    {\rm Seq} = \big[\mathbf{I}, \mathbf{t}, \mathbf{s}_0, \mathbf{o}_0, \mathbf{d}_0, ..., \mathbf{s}_t, \mathbf{o}_t, \mathbf{d}_t, ... \big],
\end{equation}
which incorporates the provided dense caption of each video clip as world dynamics $\mathbf{d}_t$.
For supervision, we employ the next-token prediction paradigm and supervise $\mathbf{d}_t$ with its textual ground truth in a teacher-forcing style.
Also, we still keep the denoising target in Eq.~(\ref{eq: denoise target}) at this stage.

\section{Experiments}
\label{sec: exps}

\begin{table*}[!ht]
\footnotesize
\centering
\caption{\textbf{Evaluation results on VBench-Long.} Subj., Bkgd., Cons., Temp., Flick., Smooth., Relation. and Appear. denote Subject, Background, Consistency, Temporal, Flickering, Smoothness, Relationship and Appearance, respectively. \textbf{Bold:} best results. \underline{Underline}: second best.
Our model achieves comparable performance with the open-sourced video generation models.
}
\vspace{-2mm}
\tabcolsep=0.15cm
\def\arraystretch{1.0}
\setlength\tabcolsep{1.1 pt}
\begin{tabular}{@{}lccccccccccccccccc@{}}
\toprule
    Method  &  \begin{tabular}[c]{@{}c@{}}Subj.\\ Cons.\end{tabular} %
    & \begin{tabular}[c]{@{}c@{}}Bkgd.\\ Cons.\end{tabular} %
    &  \begin{tabular}[c]{@{}c@{}}Temp. \\ Flick.\end{tabular} %
    & \begin{tabular}[c]{@{}c@{}}Motion\\ Smooth.\end{tabular} %
    &  \begin{tabular}[c]{@{}c@{}}Dynamic\\ Degree\end{tabular} %
    &  \begin{tabular}[c]{@{}c@{}}Aesthetic\\ Quality\end{tabular} %
    &  \begin{tabular}[c]{@{}c@{}}Imaging\\ Quality\end{tabular} %
    &  \begin{tabular}[c]{@{}c@{}}Object\\ Class\end{tabular} %
    &  \begin{tabular}[c]{@{}c@{}}Multiple\\ Objects\end{tabular} %
    &  \begin{tabular}[c]{@{}c@{}}Human\\ Action\end{tabular} %
    &  \begin{tabular}[c]{@{}c@{}}Color\end{tabular} %
    &  \begin{tabular}[c]{@{}c@{}}Spatial\\ Relation.\end{tabular} %
    &  \begin{tabular}[c]{@{}c@{}}Scene\end{tabular} %
    &  \begin{tabular}[c]{@{}c@{}}Appear.\\ Style\end{tabular} %
    &  \begin{tabular}[c]{@{}c@{}}Temp.\\ Style\end{tabular} %
    &  \begin{tabular}[c]{@{}c@{}}Overall\\ Cons.\end{tabular} %
    &  \begin{tabular}[c]{@{}c@{}}Total \\ Score\end{tabular} %
    \\ 
 \midrule
 Mira~\cite{ju2024miradata} &96.23 &96.92 &98.29 &97.54 &60.33 &42.51 &60.16 &52.06 &12.52 &63.80 &42.24 &27.83 &16.34 &21.89 &18.77 &18.72 & 71.87 \\
 
 OpenSoraPlan~\cite{opensoraplan} &95.73 &96.73 &99.03 &98.28 &47.72 &56.85 &62.28 &76.30 &40.35 &86.80 &89.19 &53.11 &27.17 &22.90 &23.87 &26.52 & 78.00 \\
 
 OpenSora~\cite{opensora}  &96.75 &\underline{97.61} &\underline{99.53} &98.50 &42.39 &56.85 &63.34 &82.22 &51.83 &91.20 &\underline{90.08} &68.56 &42.44 &23.95 &24.54 &26.85 & 79.76 \\
 
Mochi-1  &96.99 &97.28 &99.40 &99.02 &61.85 &56.94 &60.64 &86.51 &50.47 &94.60 &79.73 &69.24 &36.99 &20.33 &23.65 &25.15 & 80.13 \\

CogVideoX~\cite{yang2024cogvideox}  &96.23 &96.52 &98.66 &96.92 &\textbf{70.97} &61.98 &62.90 &85.23 &62.11 &\textbf{99.40} &82.81 &66.35 &53.20 &\textbf{24.91} &\underline{25.38} &\textbf{27.59} & 81.61 \\

Kling  &\textbf{98.33} &97.60 &99.30 &\textbf{99.40} &46.94 &61.21 &65.62 &87.24 &68.05 &93.40 &89.90 &\underline{73.03} &50.86 &19.62 &24.17 &26.42 & 81.85 \\

Vchitect-2.0~\cite{wang2023lavie}  &96.83 &96.66 &98.57 &98.98 &63.89 &60.41 &65.35 &86.61 &\underline{68.84} &\underline{97.20} &87.04 &57.55 &\textbf{56.57} &23.73 &25.01 &\underline{27.57} & 82.24 \\

Gen-3  &97.10 &96.62 &98.61 &99.23 &60.14 &\textbf{63.34} &\underline{66.82} &87.81 &53.64 &96.40 &80.90 &65.09 &\underline{54.57} &24.31 &24.71 &26.69 & \underline{82.32} \\

MiniMax  &97.51 &97.05 &99.10 &99.22 &\underline{64.91} &\underline{63.03} &\textbf{67.17} &\underline{87.83} &\textbf{76.04} &92.40 &\textbf{90.36} &\textbf{75.50} &50.68 &20.06 &\textbf{25.63} &27.10 & \textbf{83.41}\\

\midrule
\textbf{Owl-1} & \underline{98.29} & \textbf{98.61} & \textbf{99.84} & \underline{99.35} & 13.19 & 60.64 & 66.33 & \textbf{91.31} & 43.04 & 85.67 & 87.92 & 67.58 & 51.46 & \underline{24.83} & 24.25 & 25.10 & 79.65\\ \bottomrule
\end{tabular}
\label{tab:longbench}
\vspace{-5mm}
\end{table*}

\subsection{Datasets and Benchmarks}
\textbf{Geneal video generation datasets.}
We use two general purpose video generation datasets in the first two training stages.
The WebVid dataset~\cite{bain2021frozen} comprises over 10 million captioned videos sourced from the internet, totaling approximately 52K hours of footage. 
This large-scale text-video dataset encompasses a diverse range of content across multiple domains, making it highly suitable for tasks such as video-text retrieval and video generation.
We take around 400K randomly sampled videos from this dataset.
The Panda70m dataset~\cite{chen2024panda} includes 70 million videos with an average length of 8s along with their high-quality textual captions from an automatic captioning pipeline leveraging multimodal inputs and multiple cross-modal teacher models.
We randomly sample 2M videos from this dataset.

\textbf{Dense video captioning datasets.}
Due to the lack of datasets specifically focusing on the dynamics driving the progression of videos, we utilize dense video caption datasets as an alternative.
The ActivityNet Captions dataset~\cite{caba2015activitynet} contains 20K YouTube videos with 100K caption annotations and an average duration of 120 seconds.
The majority of the videos contain more than three annotated events, each associated with corresponding time span and manually written sentences, averaging 13.5 words per annotation.
The Vript dataset~\cite{yang2024vript} represents a large-scale, fine-grained video-text dataset comprising 12K high-resolution videos and over 400K segments, which are densely annotated in the form of video scripts.
The average lengths of video clips and captions are 11s and 145 words, respectively.
We use the training splits of these two datasets.

\textbf{VBench.}
VBench~\cite{huang2023vbench} is a comprehensive and hierarchical benchmark framework, which dissects video generation quality into 16 specific and disentangled dimensions, such as subject identity inconsistency, motion smoothness, temporal flickering, and spatial relationship, each equipped with tailored prompts and evaluation methodologies. 
VBench possesses three key attributes: its comprehensive coverage of diverse video generation aspects, alignment with human perception, and insights into current models' performance across various dimensions and content.

\subsection{Implementation Details}
We use the Chameleon model~\cite{team2024chameleon} as the LMM, and the DynamiCrafter-1024~\cite{xing2025dynamicrafter} as the video diffusion model.
For the trainable parameters, we finetune the LMM using LoRA~\cite{hu2021lora} and finetune all the parameters of the video diffusion model.
For the segmentation of videos, we divide each video into equal clips of 4 seconds as observations $\mathbf{o}_t$, and sample 2 frames from each clip as input to the LMM.
We set the length of learnable state queries $\mathbf{s}_t$ as 128.
For the alignment and generative pretraining stages, we train on a total of 2.4M videos from WebVid and Panda10m for 10K and 10K iterations, respectively.
For the world model training stage, we train on a total of 20K videos from ActivityNet Captions and Vript for 1K steps.

\begin{figure*}[t]
\centering
\includegraphics[width=\textwidth]{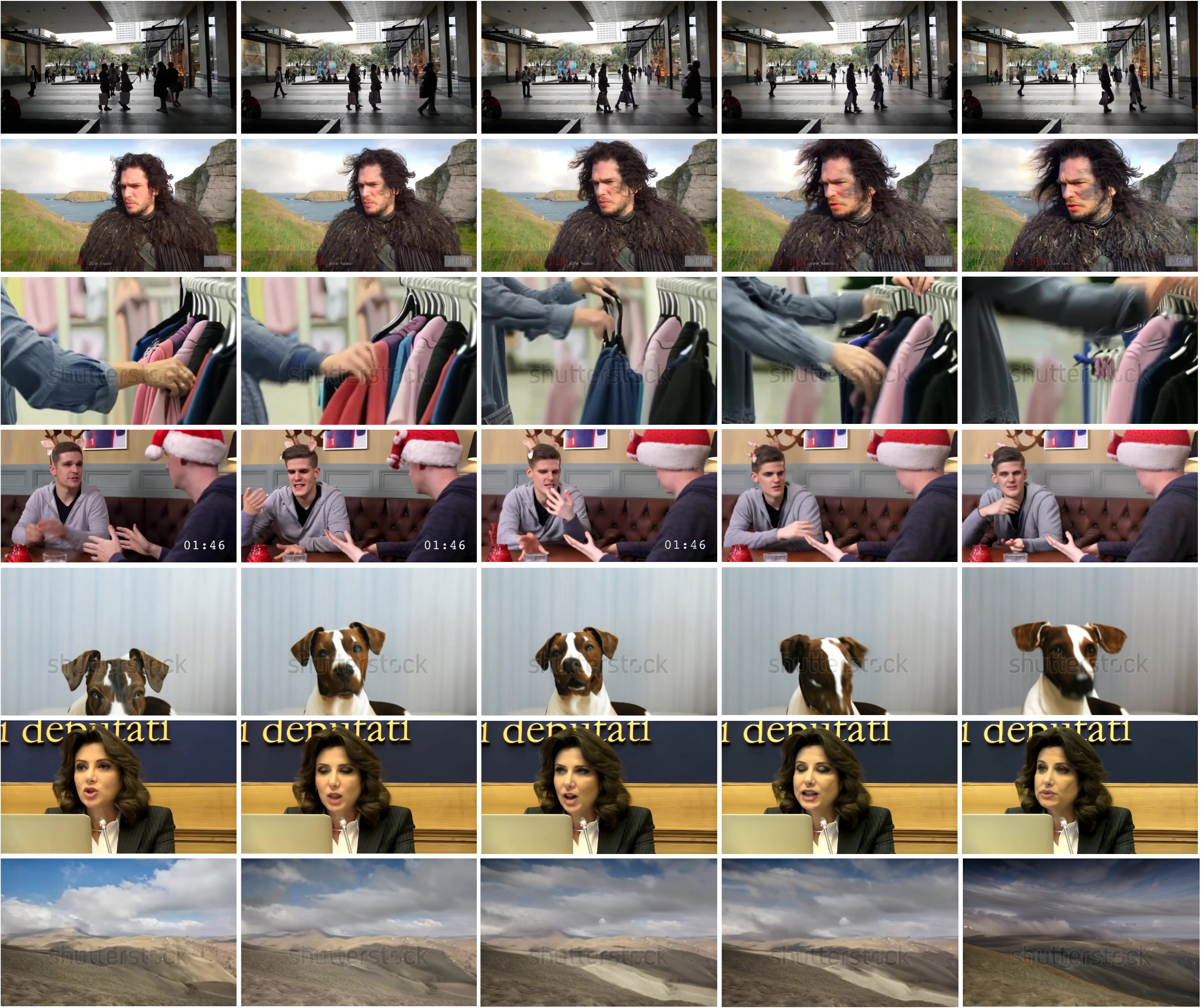}
\vspace{-5.4mm}
\caption{\textbf{Video frames visualization results for general video generation.}
We sample 5 frames from each of our generated videos, which lasts 8 seconds.
Our Owl-1 generates videos covering various topics with good quality.
}
\label{fig:short videos}
\vspace{-5.3mm}
\end{figure*}

\subsection{General Video Generation}

We evaluate our Owl-1 on two benchmarks of VBench~\cite{huang2023vbench}, i.e. VBench-I2V and VBench-Long, for its ability of generating short and long videos, respectively.
We report the results on VBench-I2V in Table~\ref{tab:i2vbench}, for which we generate 2s short videos.
Our Owl-1 achieves comparable performance with state-of-the-art methods for short video generation, excelling at the aspects of motion smoothness, background consistency and temporal flickering.
This proves the effectiveness of the state decoding mechanism which films latent state varibles into explicit video observations.
However, we do observe a decrease in the score for dynamic degree compared with DynamiCrafter, which we attribute to the lack of training video data with high motion levels.

We report the results on VBench-Long in Table~\ref{tab:longbench}, where we generate videos of 7s long, similar to the other methods.
Since the video diffusion model we use, i.e. DynamiCrafter, requires both an image and a text description as input to generate the first clip, we adopt an image diffusion model SD2.1-v~\cite{rombach2022high} to generate the first frame of the video from the given text prompt.
Our model achieves comparable performance with the open-sourced video generation models, e.g. OpenSora, on this benchmark.
Similar to the results on VBench-I2V, our Owl-1 performs better at subject and background consistency, temporal flickering and motion smoothness, while its dynamic degree is lower than other methods, which could be improved through further training with videos of higher motion level.

We visualize the generated videos of our Owl-1 in Figure~\ref{fig:short videos}.
Each of these generated videos lasts 8 seconds, and we uniformly sample 5 frames from each of them.
Owl-1 is able to generate both comprehensive and realistic videos covering various topics, including human actions, animals, natural scenery, etc.
Although we do not predict world dynamics when generating these videos, the temporal consistency remains excellent, demonstrating the effectiveness of our proposed conditional video generation approach based on state variables.
The second row of Figure~\ref{fig:short videos} captures the fine-grained detail of the face of the man, validating the ability of our model to generate high resolution videos.
\vspace{-1mm}

\begin{figure*}[t]
\centering
\includegraphics[width=\textwidth]{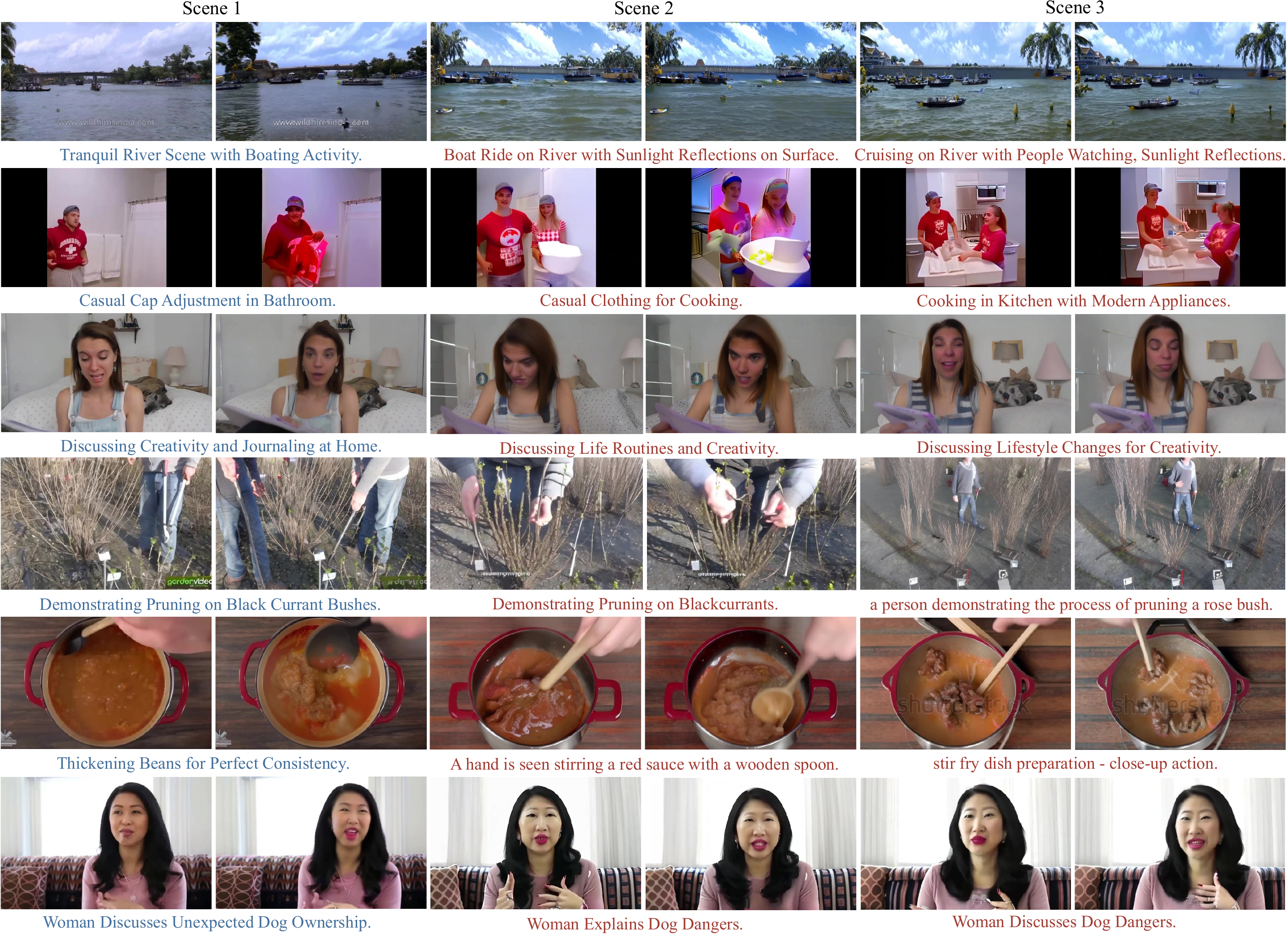}
\vspace{-6mm}
\caption{\textbf{Video frames visualization results for world model based video generation.}
We generate 3 scenes for each prompt, and sample 2 frames from each scene.
Every scene lasts for 8 seconds, and the whole video is 24 seconds long.
Our Owl-1 generates consistent long videos with reasonable dynamics anticipation.
Blue and red texts denote given prompt and predicted dynamics, respectively.
}
\label{fig:long videos}
\vspace{-6mm}
\end{figure*}

\subsection{World Model Based Video Generation}
\vspace{-1mm}
Given the current absence of benchmarks for evaluating world models in video generation, we assess the capabilities of our model through qualitative means.
We provide the visualization results of generated long videos in Figure~\ref{fig:long videos}.
We generate 3 scenes for a given prompt, and sample 2 frames from each scene. 
Every scene lasts for 8 seconds, and the whole video is 24 seconds long. 
When transitioning from one scene to another, we manually discard the image condition from the last frame and depend solely on the latent state variable as condition for geneartion, which is challenging because the latent state has to include information about the style and context of the previous video clips to generate the next clip in a consistent manner.
We observe that Owl-1 is able to generate consistent long videos with reasonable dynamics anticipation.
The video in the fourth row features a man engaged in gardening, where he utilizes tools to prune branches.
The video we generated initially focuses on his hand movements and subsequently showcases the overall pruning effect, demonstrating a certain degree of logic. 
This reflects the modeling and prediction of the evolution of the world. 
However, we do notice that the predicted dynamics exhibit a certain degree of repetition, which we hypothesize is due to the inherent repetitiveness in the dense captions of the training video data.
Even so, the videos generated by Owl-1 maintain good consistency across different scenes.

\vspace{-3mm}
\section{Conclusion}
\label{sec: conclusion}
\vspace{-1mm}
In this paper, we have proposed an Omni World Model (Owl-1) for consistent long video generation.
Our Owl-1 approaches this task from the perspective of world model, which models the evolution of the world with a sequence of state variables. %
We have introduced a closed-loop state-observation-dynamics triplet, in which the latent states encode both current and historical information about the world and serve as comprehensive long-horizon conditions for video generation.
Explicit video observations are then decoded from latent state variables with a video diffusion model.
To drive the world evolution, we incorporate the anticipation of the world dynamics during the generation process, which is beneficial for the diversity of generated content.
Furthermore, we have devised an effective mutli-stage training scheme for our Owl-1 to take advantage of the vast amount of short video data and only finetune on a relatively small amount of long video data which reflects the evolution of the world.
Owl-1 shows impressive capabilities in generating long and consistent videos.
The visualizations further validate Owl-1's ability to capture fine-grained details and generate videos with reasonable dynamics anticipation.

\textbf{Limitations and future work.}
From the evaluation and visualization results, we do notice some limitations of the current Owl-1, especially the decreased dynamic degree after fintuning the video diffusion model and repetitive world dynamics.
Future work could investigate into these drawbacks and the scale up of our model on a large amount of high-quality video data with dense captions featuring the evolution process of the world.
We believe the proposed paradigm for video generation world model is one of the approaches to realize multimodal general intelligence.

\appendix

\begin{figure*}
    \centering
    \includegraphics[width=\linewidth]{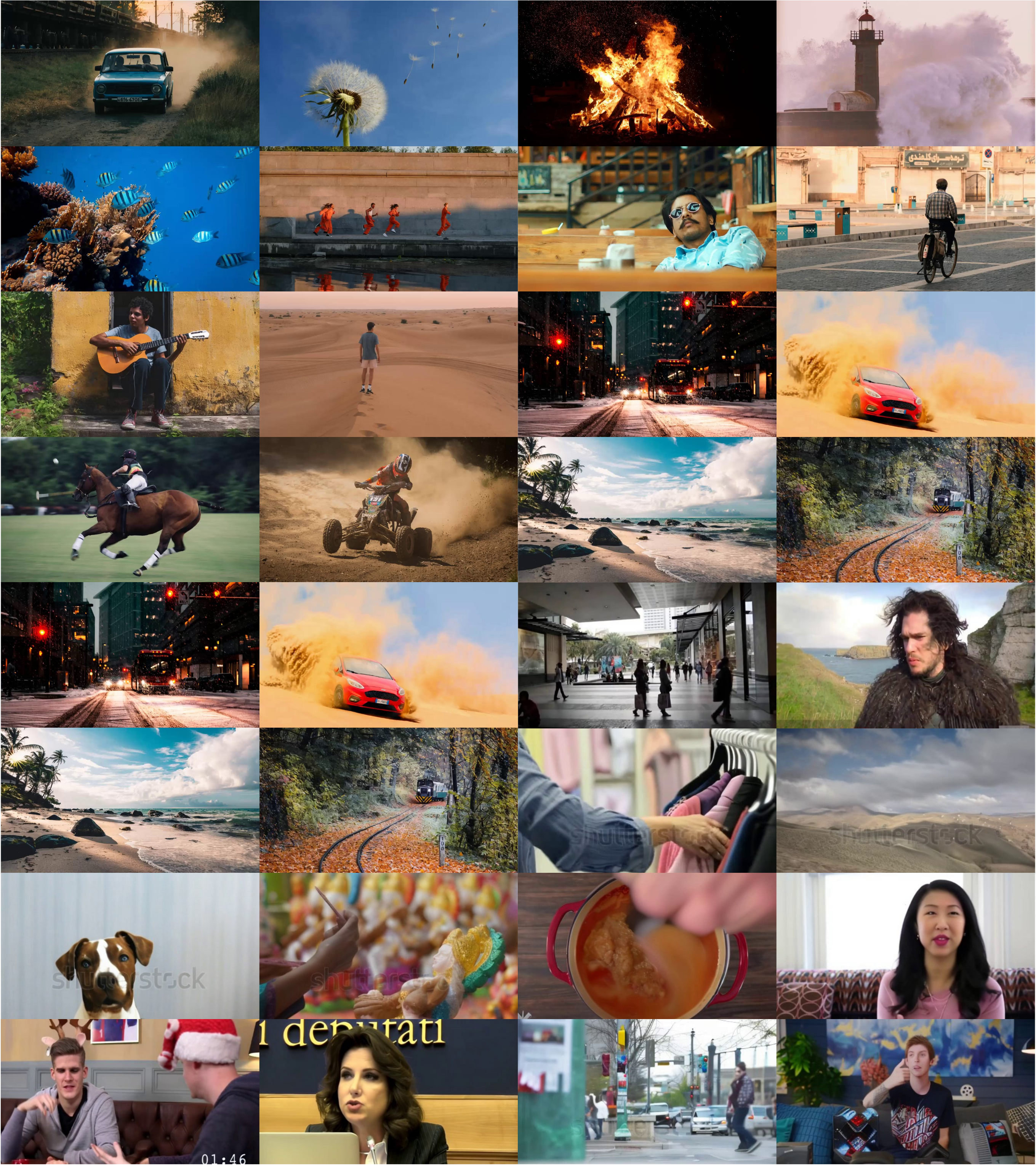}
    \vspace{-5mm}
    \captionof{figure}{
    \textbf{Gallery of various video samples of our Owl-1.}
    We take one frame from each of these samples for demonstration.
    }
\label{fig: supp teaser}
\vspace{-5mm}
\end{figure*}%

\section{Video Samples}
\label{sec:supp videos}
We provide more samples\footnote{\url{https://github.com/huang-yh/Owl}} generated by our Owl-1 based on a wider range of prompts in Figure~\ref{fig: supp teaser}.
In addition to the samples shown in the main paper which use the video frames from the validation or test sets of our training datasets as prompts, we also visualize some samples generated according to the standard prompts from the VBench-I2V benchmark~\cite{huang2023vbench} with higher quality.
As shown by Figure~\ref{fig: supp teaser}, our model is able to generate videos covering a variety of topics, and the quality of generated videos generally improves given image prompts of higher quality.

\section{Additional Implementation Details}
When finetuning Chameleon~\cite{team2024chameleon} as the LMM, We employ LoRA~\cite{hu2021lora} and set the rank of LoRA to 8, resulting in approximately 798M trainable parameters.
Together with all the parameters from DynamiCrafter~\cite{xing2025dynamicrafter} as the video diffusion model, the total amount of trainable parameters is about 2B.
We train the Owl-1 using 8 A800 GPUs with 80G memory, and the training time for the three stages is 1 day, 5 days, and 1 day, respectively.

\section{Controllability over Scene Transitions}
Due to the scarcity of high quality video data with varying temporal content and devoid of scene transitions, we adopt the datasets of dense video captioning as the training data for the world model training stage.
However, these datasets, e.g. Vript~\cite{yang2024vript} and ActivityNet Captions~\cite{caba2015activitynet}, often incorporate scene transitions in a long video, which poses challenge for the training process.
To address this issue, we manually discard the concatenating image conditions when generating the next clip belonging to a new scene during training.
This strategy also endows our model with the capability to perform controllable scene transitions.
Similar to the training phase, we only need to omit the concatenating image conditions to transit into a new scene.
When generating longer videos in Figure~\ref{fig: supp teaser} and Fig.~4 in the main paper, we set the interval between scene transitions to about 2 short clips generated by the video diffusion model, resulting in the duration of each scene being about 4 seconds.

\end{document}